\documentclass[10pt,twocolumn,letterpaper]{article}

\usepackage{cvpr}
\usepackage{times}
\usepackage{epsfig}
\usepackage{graphicx}
\usepackage{amsmath}
\usepackage{amssymb}
\usepackage{color} 
\usepackage{cite} 
\usepackage{booktabs} 
\usepackage{enumitem}
\usepackage{authblk}

\definecolor{julieta_color}{RGB}{56,108,176} 
\definecolor{javier_color}{RGB}{217,95,2}    
\definecolor{michael_color}{RGB}{27,158,119} 

\graphicspath{ {./imgs/} }

\usepackage[pagebackref=true,breaklinks=true,letterpaper=true,colorlinks,bookmarks=false]{hyperref}

\cvprfinalcopy 

\def\cvprPaperID{****}

\ifcvprfinal\fi
\begin{document}

\title{On human motion prediction using recurrent neural networks\vspace{-2ex}}

\author[1]{Julieta Martinez\thanks{Research carried out while Julieta was an intern at MPI.}}
\author[2]{Michael J. Black}
\author[3]{Javier Romero\vspace{-1ex}}
\affil[1]{University of British Columbia, Vancouver, Canada}
\affil[2]{MPI for Intelligent Systems, T{\"u}bingen, Germany}
\affil[3]{Body Labs Inc., New York, NY\vspace{0.5ex}}
\affil[ ]{\tt\small julm@cs.ubc.ca, black@tuebingen.mpg.de, javier.romero@bodylabs.com\vspace{-2ex}}
\maketitle

\begin{abstract}
Human motion modelling is a classical problem at the intersection of graphics and computer vision, with applications spanning human-computer interaction, motion synthesis, and motion prediction for virtual and augmented reality.
Following the success of deep learning methods in several computer vision tasks, recent work has focused on using deep recurrent neural networks (RNNs) to model human motion, with the goal of learning time-dependent representations that perform tasks such as short-term motion prediction and long-term human motion synthesis. We examine recent work, with a focus on the evaluation methodologies commonly used in the literature, and show that, surprisingly, state-of-the-art performance can be achieved by a simple baseline that does not attempt to model motion at all. We investigate this result, and analyze recent RNN methods by looking at the architectures, loss functions, and training procedures used in state-of-the-art approaches. We propose three changes to the standard RNN models typically used for human motion, which result in a simple and scalable RNN architecture that obtains state-of-the-art performance on human motion prediction.
\end{abstract}

\begin{figure}
  \includegraphics[width=\linewidth,trim=104mm 56mm 105mm 20mm,clip=true]{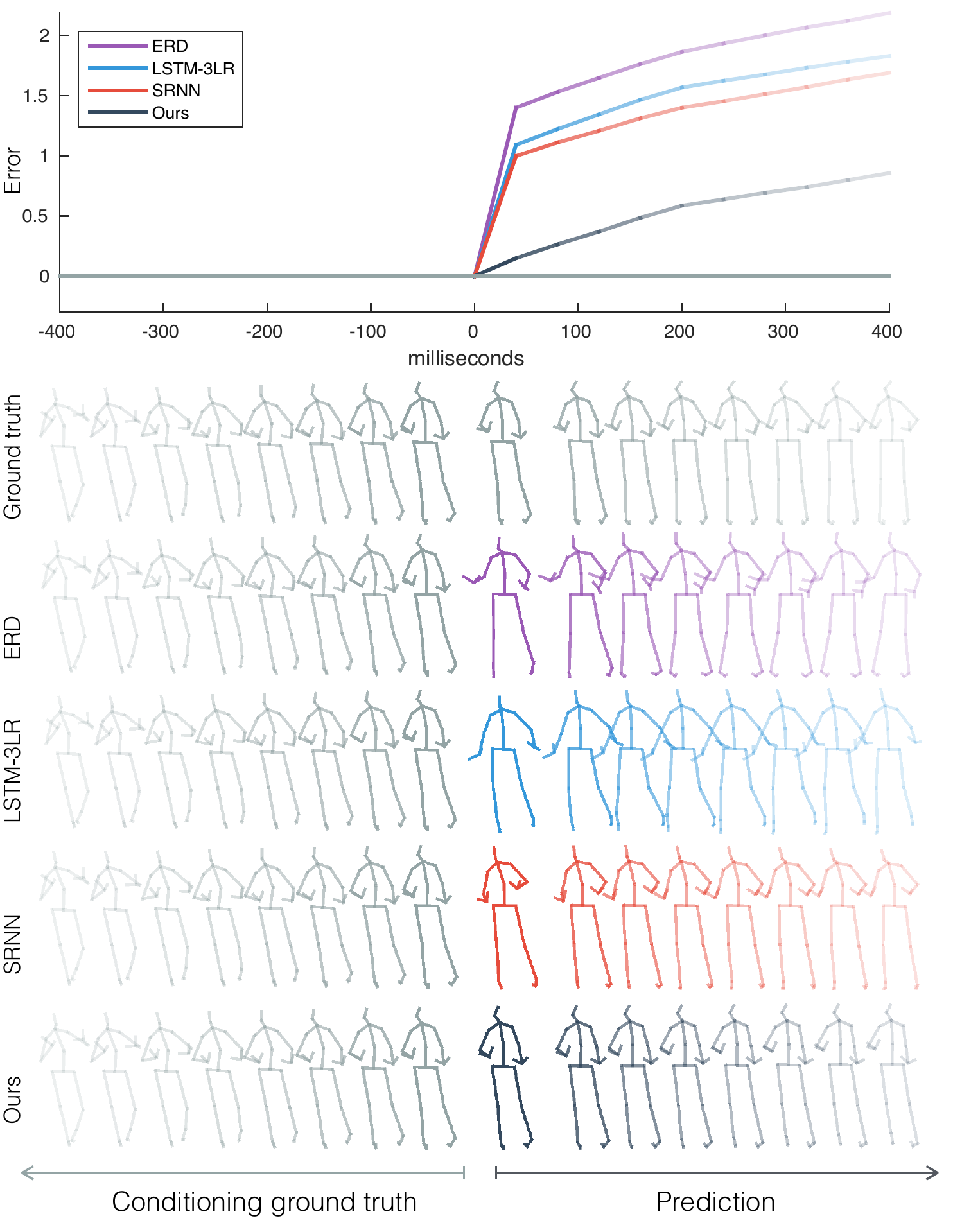}
  \caption{Top: Mean average prediction error for different motion prediction methods. Bottom: Ground truth passed to the network is shown in grey, and short-term motion predictions are shown in colour. Previous work, based on deep RNNs, produces strong discontinuities at the start of the prediction (middle column). Our method produces smooth, low-error predictions.}
  \label{fig:teaser}
  \vspace{-2mm}
\end{figure}

\vspace{-2mm}
\section{Introduction}

An important component of our capacity to interact with the world resides in the ability to predict its evolution over time.
Handing over an object to another person, playing sports, or simply walking in a crowded street would be extremely challenging tasks without our understanding of how people move, and our ability to predict what they are likely to do in the following instants.
Similarly, machines that are able to perceive and interact with moving people, either in physical or virtual environments,
must have a notion of how people move.
Since human motion is the result of both physical limitations (\eg torque exerted by muscles, gravity, moment preservation) and the intentions of subjects (\emph{how} to perform an intentional motion), motion modeling is a
complex task that 
should be ideally learned from observations.

Our focus in this paper is to learn models of human motion from motion capture (mocap) data.
More specifically, we are interested in human motion prediction, where we forecast the most likely future 3D poses
of a person given their past motion.
This problem has received interest in a wide variety of fields, such as action prediction for socially-aware
robotics~\cite{koppula}, 3D people tracking within computer vision~\cite{3dtracking, gupta20143dpose},
motion generation for computer graphics~\cite{motion-graphs} or modeling biological motion in
psychology~\cite{troje:2002}. 

Traditional approaches have typically imposed expert knowledge about motion in their systems in the form of Markovian
assumptions~\cite{pavlovic:2000,lehrmann-a-non-parametric}, smoothness, or low dimensional embeddings~\cite{gpdm}.
Recently, a family of methods based on deep recurrent neural networks (RNNs) have shown good performance on this task
while trying to be more agnostic in their assumptions.
For example, \cite{fragkiadaki} uses curriculum learning and incorporates representation learning in the architecture,
and~\cite{srnn} manually encodes the semantic similarity between different body parts.
These approaches benefit from large, publicly available collections of motion capture data~\cite{h36m},
as well as recent advances in the optimization of time-series modelling~\cite{gru}.

Recent work has validated its performance via two complementary methods: 
(1) \emph{quantitative} prediction error in the short-term, typically measured as a mean-squared loss in angle-space,
and (2) \emph{qualitative} motion synthesis for longer time horizons, where the goal is to generate feasible motion.
The first evaluation metric is particularly interesting for computer vision applications such as people tracking,
where predictions are continually matched and corrected with new visual evidence.
The second criterion, most relevant for open-loop motion generation in graphics, is hard to evaluate
quantitatively, because human motion is a highly non-deterministic process over long time horizons.
This problem is similar to the one found in recent research on deep generative networks~\cite{on-the-evaluation-of-gen-models},
where the numerical evaluation based on the negative log-likelihood and Parzen window estimates are known to be far from perfect.

We have empirically observed that current deep RNN-based methods have difficulty obtaining good performance on both tasks.
Current algorithms are often trained to minimize a quantitative loss for short-term prediction, while striving to
achieve long-term plausible motion by tweaking the architectures or learning procedures. As a result, their long-term results suffer from occasional
unrealistic artifacts such as foot sliding, while their short-term results are not practical for tracking due to clear
discontinuities in the first prediction. In fact, the discontinuity problem is so severe,
that we have found that state-of-the-art methods are quantitatively outperformed by a range of simple baselines,
including a constant pose predictor. While this baseline does not produce interesting motion in the long-run,
it highlights both a poor short-term performance, as well as a severe discontinuity problem in current deep RNN approaches.
In this work, we argue that (a) the results achieved by recent work are not fully satisfactory
for either of these problems, and (b) trying to address both problems at once is very challenging,
especially in the absence of a proper quantitative evaluation for long-term plausibility. 

We focus on short-term prediction, which is the most relevant task for a visual tracking scenario.
We investigate the reasons for the poor performance of recent methods on this task by analyzing several factors such as
the network architectures and the training procedures used in state-of-the-art RNN methods.
First, we consider the training schedule used in~\cite{fragkiadaki, srnn}.
It is a known problem in RNNs~\cite{scheduled-sampling} and reinforcement learning~\cite{dagger} that networks cannot learn
to recover from their own mistakes if they are fed only ground-truth during training.
The authors of~\cite{fragkiadaki, srnn} introduced increasing amounts of random noise during training to compensate for
this effect. However, this noise is difficult to tune, makes it harder to choose the best model based on validation error,
and has the effect of degrading the quality of the prediction in the first frame.
Instead, we propose a simple approach that introduces realistic error in training time without any scheduling;
we simply feed the predictions of the net, as it is done in test time. This increases the robustness of the predictions
compared to a network trained only on ground truth, while avoiding the need of a difficult-to-tune schedule.

Unfortunately, this new architecture is still unable to accurately represent the conditioning poses in its hidden representation, which still results 
in a discontinuity in the first frame of the prediction. We borrow ideas from research on the statistics of hand motion~\cite{Ingram2008}, and model velocities instead of absolute joint angles, while keeping the loss
in the original angle representation to avoid drift.
Therefore, we propose a \emph{residual} architecture that models first-order motion derivatives, which results in smooth and much more accurate short-term predictions.

Both of our contributions can be implemented using an architecture that is significantly simpler than those in previous work.
In particular, we move from the usual multi-layer LSTM architectures (long short-term memory) to a single GRU (Gated Recurrent Unit), and do not require a spatial encoding layer. This allows us to train a single model on the entire Human 3.6M dataset~\cite{h36m} in a few hours.
This differs from previous approaches~\cite{fragkiadaki,srnn}, which trained only action-specific models from
that dataset. Our approach sets the new state of the art on short-term motion prediction,
and overall gives insights into the challenges of motion modelling using RNNs.
Our code is publicly available 
at \url{https://github.com/una-dinosauria/human-motion-prediction}.

\section{Related work}

Our main task of interest is human motion prediction, with a focus on recent deep RNN architectures~\cite{fragkiadaki,srnn}. One of our findings is that, similar to~\cite{revisiting-vqa, arora}, a family of simple baselines outperform recent deep learning approaches. We briefly review the literature on these topics below.

\vspace{-3mm}
\paragraph{Modelling of human motion.}

Learning statistical models of human motion is a difficult task due to the high-dimensionality, non-linear dynamics and stochastic nature of human movement. Over the last decade, and exploiting the latent low-dimensionality of action-specific human motion, most work has focused on extensions to latent-variable models that follow state-space equations such as hidden Markov models (HMMs)~\cite{lehrmann-efficient-non-linear}, exploring the trade-offs between model capacity and inference complexity. For example, Wang~\etal~\cite{gpdm} use Gaussian-Processes to perform non-linear motion prediction, and learn temporal dynamics using expectation maximization and Markov-chain Monte Carlo. Taylor~\etal~\cite{crbm} assume a binary latent space and model motion using a conditional restricted Boltzman machine (CRBM), which requires sampling for inference. Finally, Lehrmann~\etal~\cite{lehrmann-efficient-non-linear} use a random forest to non-linearly choose a linear system that predicts the next frame based on the last few observations.

\vspace{-3mm}
\paragraph{Applications of human motion models.}
Motion is a key part of actions; therefore, the field of action recognition has paid special attention to models and representations of human motion.
In their seminal work, Yacoob and Black~\cite{Yacoob:1998:PMR} model motion with time-scale and time-shifted activity bases from a linear manifold of visual features computed with Principal Component Analysis (PCA). More complex models like mixtures of HMMs~\cite{Raman2015149, LoPresti201529}, latent topic models of visual words~\cite{mori:PAMI:2007} or LSTMs~\cite{Liu2016} are used in recent methods. Although their purpose (action classification from a sequence of poses) is different from ours, this field contains interesting insights for motion prediction, such as the importance of a mathematically sound orientation representation~\cite{Vemulapalli_2014_CVPR, Vemulapalli_2016_CVPR} or how learned, compact motion representations improve action recognition accuracy~\cite{Mahasseni_2016_CVPR}.

Another popular use of motion models, specially short-term ones, is pose tracking. The use of simple linear Markovian models~\cite{Sidenbladh:2000:STH} or PCA models~\cite{Urtasun:2006:Temp} has evolved to  locally linear ones like factor analizers~\cite{li:2006:ECCV}, non-linear embeddings like Laplacian Eigenmaps~\cite{Sminchisescu:2004:GMC}, Isomap~\cite{Jenkins:2004:SEI}, dynamic variants of Gaussian Process Latent Variable Models (GPLVM)~\cite{gpdm,conf/nips/YaoGGU11}, or physics-based models~\cite{journals/ijcv/BrubakerFH10}.

In animation, similar methods have been used for the generation of human pose sequences. Spaces of HMMs parameterised by style were used by Brand~\etal~\cite{Brand:2000:SM} to generate complex motions. Arikan and Forsyth~\cite{journals/tog/ArikanF02} collapse full sequences into nodes in a directed graph, connected with possible transitions between them, and in~\cite{journals/tog/LeeCRHP02} cluster trees improve the path availability. More recently, motion models based on GPLVM have been used for controlling virtual characters in a physical simulator~\cite{levine:TOG:2012}. An overview of motion generation for virtual characters can be found in ~\cite{welbergen:CGF:2010}.

\vspace{-3mm}
\paragraph{Deep RNNs for human motion.}
Our work focuses on recent approaches to motion modelling that are based on deep RNNs. 
Fragkiadaki~\etal~\cite{fragkiadaki} propose two architectures: LSTM-3LR (3 layers of Long Short-Term Memory cells) and ERD (Encoder-Recurrent-Decoder). Both are based on concatenated LSTM units, but the latter adds non-linear space encoders for data pre-processing. The authors also note that, during inference, the network is prone to accumulate errors, and quickly produces unrealistic human motion. Therefore, they propose to gradually add noise to the input during training (as is common in curriculum learning~\cite{curriculum-learning}), which forces the network to be more robust to prediction errors. This noise scheduling makes the network able to generate plausible motion for longer time horizons, specially on cyclic walking sequences. However, tuning the noise schedule is hard in practice.

More recently, Jain~\etal~\cite{srnn} introduced structural RNNs (SRNNs), an approach that takes a manually designed graph that encodes semantic knowledge about the RNN as input, and creates a bi-layer architecture that assigns individual RNN units to semantically similar parts of the data. The authors also employ the noise scheduling technique introduced by Fragkiadaki~\etal, and demonstrate that their network outperforms previous work both quantitatively in short-term prediction, as well as qualitatively. Interestingly, SRNNs produce plausible long-term motion for more challenging, locally-periodic actions such as eating and smoking,
and does not collapse to unrealistic poses in aperiodic ``discussion'' sequences.

\vspace{-3mm}
\paragraph{Revisiting baselines amid deep learning.}
The rise and impressive performance of deep learning methods in classical problems such as object recognition~\cite{alexnet}
has encouraged researchers to attack both new and historically challenging problems using variations of deep neural networks. For example, there is a now a large body of work on visual question answering (VQA), \ie the task of answering natural-language questions by looking at images, based almost exclusively on end-to-end trainable systems with deep CNNs for visual processing and deep RNNs for language modelling~\cite{ask-attend-answer, exploring-models, ask-your-neurons}.
Recently, however, Zhou~\etal~\cite{simple-baseline} have shown that a simple baseline that concatenates features from questions' words and CNN image features performs comparably to approaches based on deep RNNs. Moreover, Jabri~\etal~\cite{revisiting-vqa} have shown competitive performance on VQA with a simple baseline that does not take images into account, and state-of-the-art performance with a baseline that is trained to exploit the correlations between questions, images and answers.

Our work is somewhat similar to that of Jabri~\etal~\cite{revisiting-vqa}, in that we have found a very simple baseline that outperforms sophisticated state-of-the-art methods based on deep RNNs for short-term motion prediction. In particular, our baseline outperforms the ERD and LSTM-3LR models by Fragkiadi~\etal~\cite{fragkiadaki}, as well as the structural RNN (SRNN) method of Jain~\etal~\cite{srnn}. Another example of baselines outperforming recent work in the field of pose models can be found in~\cite{lehrmann-a-non-parametric}, where a Gaussian pose prior outperforms the more complicated GPLVM.

\section{Method}

Recent deep learning methods for human pose prediction~\cite{fragkiadaki,srnn} offer an agnostic learning framework that could potentially be integrated with video data~\cite{fragkiadaki} or used for other forecasting applications~\cite{srnn}. However, for the specific task of motion forecasting, we note that they have a few common pitfalls that we would like to improve.
\subsection{Problems}

\paragraph{First frame discontinuity.} While both methods generate continuous motion, a noticeable jump between the conditioning ground truth and the first predicted frame is present in their results (see Figure~\ref{fig:teaser}). This jump is particularly harmful for tracking applications, where short-term predictions are continuously updated with new visual evidence.

\vspace{-3mm}
\paragraph{Hyper-parameter tuning.} These methods add to the typical set of network hyper-parameters an additional one, particularly hard to tune: the noise schedule.

In time series modelling, it is often necessary to model noise as part of the input, in order to improve robustness against noisy observations. For example, in Kalman filtering, a small amount of Gaussian noise is modelled explicitly as part of the standard state-space equations.
In applications such as motion synthesis, exposing the method to the errors that the network will make at test time is crucial to prevent the predicted poses from leaving the manifold of plausible human motion. Algorithms like DAGGER~\cite{dagger}, used in reinforcement learning, use queries to an ``expert'' during training so that the predictor learns how to correct its own errors. It is, however, not straightforward how one would use this approach for pose prediction.

\begin{figure*}
  \includegraphics[width=\linewidth,trim=4mm 65mm 4mm 80mm,clip=true]{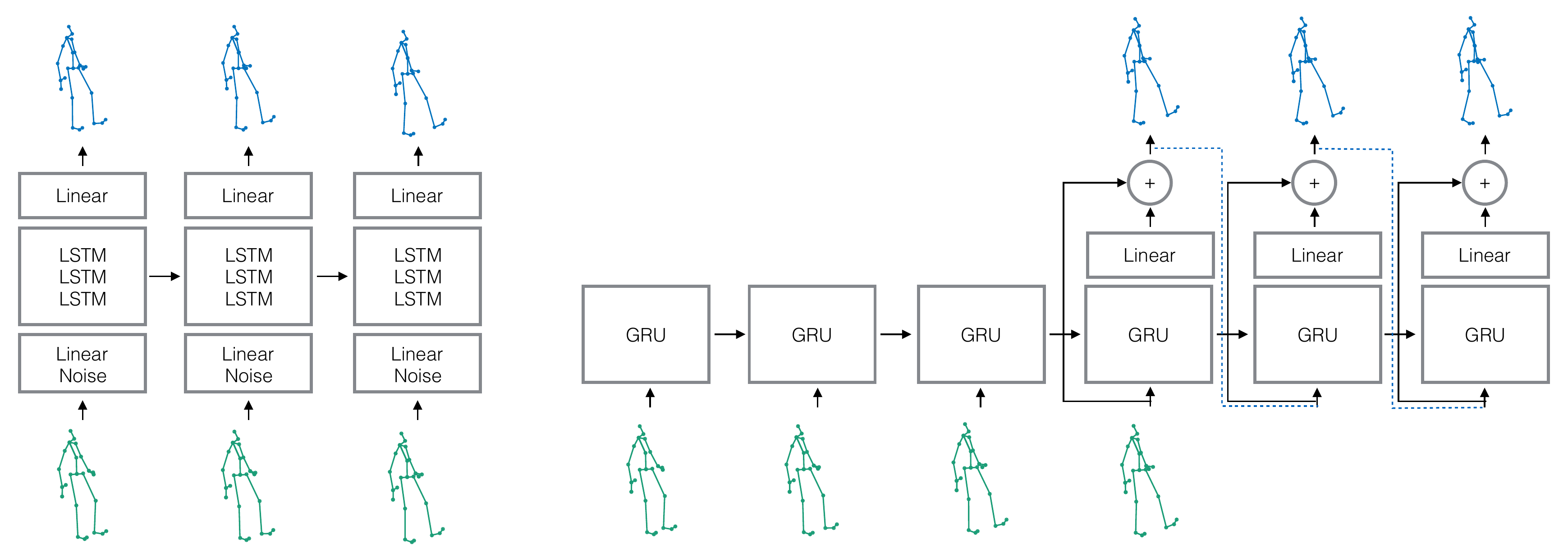}
  \caption{Training procedure as done in previous work, and our proposed sequence-to-sequence residual architecture. Green stick figures represent ground truth, and blue stick figures represent predictions. Left: LSTM-3LR architecture, introduced by Fragkiadaki~\etal~\cite{fragkiadaki}. During training, ground truth is fed to the network at each time-step, and noise is added to the input. Right: Our sequence-to-sequence architecture; During training, the ground truth is fed to an \emph{encoder} network, and the error is computed on a \emph{decoder} network that feeds its own predictions. The decoder also has a residual connection, which effectively forces the RNN to internally model angle velocities.}
  \label{fig:method}
  \vspace{-2mm}
\end{figure*}

The basic architectures that we use, RNNs, typically do not consider this mismatch between train and test input, which makes them prone to accumulate errors at inference time. To alleviate this problem, Fragkiadaki~\etal propose to use noise scheduling; that is, to inject noise of gradually increasing magnitude to the input in training time (see Fig.~\ref{fig:method}, left), which corresponds to a type of curriculum learning. Jain~\etal~\cite{srnn} similarly adopt this idea, and have found that it helps stabilizing long-term motion synthesis. The downsides are, (1) that both noise distribution and magnitude scheduling are hard to tune, (2) that while this noise improves long-term predictions, it tends to hurt performance in short-term predictions, as they become discontinuous from previous observations, and (3) that the common rule for choosing the best model, based on lowest validation error, is not valid anymore, since lowest validation error typically corresponds to the validation epoch without injected noise.

\vspace{-3mm}
\paragraph{Depth and complexity of the models.}
LSTM-3LR, ERD and SRNN use more than one RNN in their architectures, stacking two or three layers for increased model capacity. While deeper models have empirically shown the best performance on a series of tasks such as machine translation~\cite{seq2seq}, deep networks are known to be hard to train when data is scarce (which is the data regime for action-specific motion models). Moreover, recent work has shown that shallow RNNs with minimal representation processing can achieve very competitive results in tasks such as the learning of sentence-level embeddings~\cite{skip-thought}, as long as a large corpus of data is available. Finally, deeper models are computationally expensive, which is an important factor to consider in the context of large-scale training datasets.

\vspace{-3mm}
\paragraph{Action-specific networks.} Although the vision community has recently benefited from large-scale, publicly available datasets of motion capture data~\cite{h36m}, motion modelling systems have been typically trained on small action-specific subsets. While restricting the training data to coherent subsets makes modelling easier, it is also well-known that deep networks work best when exposed to large and diverse training datasets~\cite{alexnet, skip-thought}. This should specially apply to datasets like Human3.6M, where different actions contain large portions of very similar data (\eg sitting or walking).

\subsection{Solutions}

\paragraph{\bf Sequence-to-sequence~\cite{seq2seq} architecture.}
We address short-term motion prediction as the search for a function that maps an input sequence (conditioning ground truth) to an output (predicted) sequence.
In this sense, the problem is analogous to machine translation, were sequence-to-sequence (seq2seq) architectures are popular. In seq2seq, two networks are trained; (a) an encoder that receives the inputs and generates an internal representation, and (b), a decoder network, that takes the internal state and produces a maximum likelihood estimate for prediction.
Unlike the common practice in machine translation, we
enforce the encoder and the decoder to share weights, which we found to accelerate convergence. 
A benefit of this architecture is that the encoding-decoding procedure during training is more similar to the protocol used at test-time.
Moreover, there are multiple variations of seq2seq architectures (\eg, with attention mechanisms~\cite{attention}, or bi-directional encoders~\cite{exploring-models}), that could potentially improve motion prediction.

\vspace{-3mm}
\paragraph{\bf Sampling-based loss.}
While it is often common in RNNs to feed the ground truth at each training time-step to the network, this approach has the downside of the network not being able to recover from its own mistakes. Previous work has addressed this problem by scheduling the rate at which the network sees either the ground truth or its own predictions~\cite{scheduled-sampling}, or by co-training and adversarial network to force the internal states of the RNN to be similar during train and test time~\cite{professor-forcing}. These approaches, however, rely heavily on hyper-parameter tuning, which we want to avoid. Striving for simplicity, during training we let the decoder produce a sequence by always taking as input its own samples. This approach requires absolutely no parameter tuning.
Another benefit of this approach is that we can directly control the length of the sequences that we train on. As we will see, training to minimize the error on long-term motions results in networks that produce plausible motion in the long run, while training to minimize error the short-term reduces the error rate in the first few predicted frames.

\vspace{-3mm}
\paragraph{\bf Residual architecture.}
While using a seq2seq architecture trained with a sampling-based loss can produce plausible long-term motion, we have observed that there is still a strong discontinuity between the conditioning sequence and prediction. Our main insight is that motion continuity, a known property of human motion, is easier to express in terms of velocities than in poses. While it takes considerable modelling effort to represent all possible conditioning poses so that the first frame prediction is continuous, it only requires modeling one particular velocity (zero, or close to zero velocity) to achieve the same effect. This idea is simple to implement in current deep learning architectures since it translates into adding a residual connection between the input and the output of each RNN cell (see Fig.~\ref{fig:method}, right).
We note that, although residual connections have been shown to improve performance on very deep convolutional networks~\cite{residual-learning}, in our case they help us model prior knowledge about the statistics of human motion.

\vspace{-3mm}
\paragraph{\bf Multi-action models.}
We also explore training a single model to predict motion for multiple actions, in contrast to previous work~\cite{fragkiadaki, srnn}, which has focused on building action-specific models.
While modelling multiple actions is a more difficult task than modelling single-action sets, 
it is now a common practice to train a single, conditional model, on multiple data modalities, as this allows the network to exploit regularities in large datasets~\cite{graves-generating-sequences}.
Semantic knowledge about each activity can be easily incorporated using one-hot vectors; \ie, concatenating, in the input, a 15-dimensional vector that has zeros everywhere, but a value of one in the index of the indicated action.

\section{Experimental setup}

We consider three main sets of experiments to quantify the impact of our contributions:

\vspace{-3mm}
\paragraph{1. Seq2seq architecture and sampling-based loss.} First, we train action-specific models using our proposed sequence-to-sequence architecture with sampling-based loss, and compare it to previous work, which uses noise scheduling, and to a baseline that feeds the ground truth at each time-step. The goal of these experiments is to verify that using a sampling-based loss, which does not require parameter tuning, performs on par with previous work on short-term motion prediction, while still producing plausible long-term motion. In these experiments, the network is trained to minimize the loss over 1 second of motion.
  
\vspace{-3mm}
\paragraph{2. Residual architecture.} The second set of experiments explore the effects of using a residual architecture that models first-order motion derivatives, while keeping the loss in the original angle space. Here, we are interested in learning whether a residual architecture improves short term prediction; therefore, in these experiments, the network is trained to minimize the prediction error over 400 milliseconds.
  
\vspace{-3mm}
\paragraph{3. Multi-action models.} Our last round of experiments quantifies the benefits of training our architecture on the entire Human 3.6M dataset, as opposed to building action-specific models. We consider both a supervised and an unsupervised variant. The \emph{supervised} variant enhances the input to the model by concatenating one-hot vectors with the 15 action classes. In contrast, the \emph{unsupervised} variant does not use one-hot input during training nor prediction. In these experiments we also train the network to minimize the prediction error over the next 400 milliseconds.

\begin{figure}
  \includegraphics[width=\linewidth,trim=5mm 5mm 5mm 2mm,clip=true]{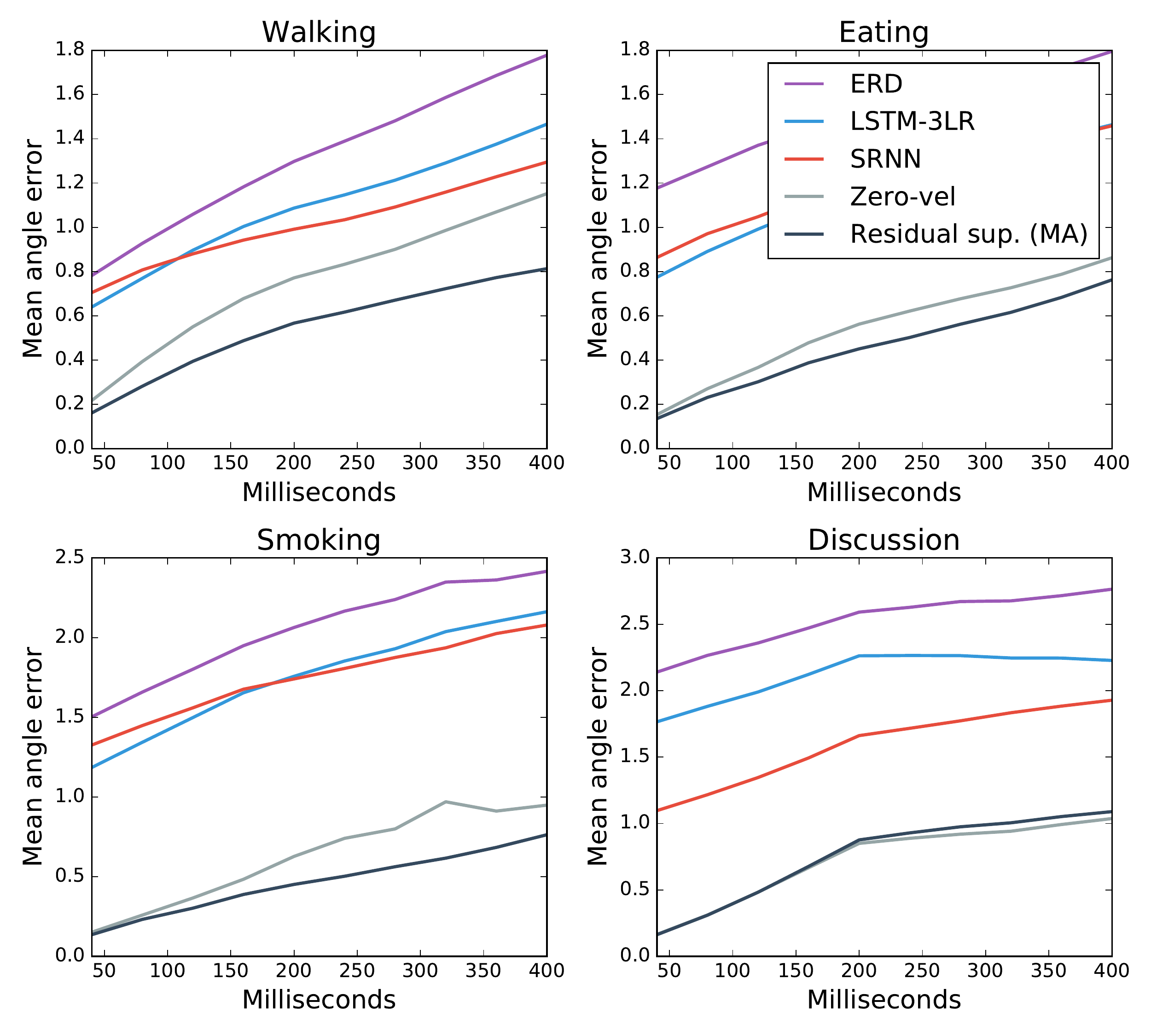}
  \caption{Error curves comparing ERD~\cite{fragkiadaki}, LSTM-3LR~\cite{fragkiadaki}, SRNN~\cite{srnn} and our method ({\bf Residual sup. (MA)} in Table~\ref{tab:detailed}) with residual connections, sampling-based loss and trained on multiple actions, as well as a zero-velocity baseline.}
  \label{fig:losses}
  \vspace{-3mm}
\end{figure}

\begin{table*}[t!]
\centering
\small
\tabcolsep=1.5mm
\begin{tabular}{@{}lrrrr|rrrr|rrrr|rrrr@{}}
 & \multicolumn{4}{c}{Walking} & \multicolumn{4}{c}{Eating} & \multicolumn{4}{c}{Smoking} & \multicolumn{4}{c}{Discussion}\\
milliseconds & 80 & 160 & 320 & 400 & 80 & 160 & 320 & 400 & 80 & 160 & 320 & 400 & 80 & 160 & 320 & 400 \\
\midrule
ERD~\cite{fragkiadaki}       & 0.93 & 1.18 & 1.59 & 1.78 & 1.27 & 1.45 & 1.66 & 1.80 & 1.66 & 1.95 & 2.35 & 2.42 & 2.27 & 2.47 & 2.68 & 2.76\\
LSTM-3LR~\cite{fragkiadaki}  & 0.77 & 1.00 & 1.29 & 1.47 & 0.89 & 1.09 & 1.35 & 1.46 & 1.34 & 1.65 & 2.04 & 2.16 & 1.88 & 2.12 & 2.25 & 2.23\\
SRNN~\cite{srnn}             & 0.81 & 0.94 & 1.16 & 1.30 & 0.97 & 1.14 & 1.35 & 1.46 & 1.45 & 1.68 & 1.94 & 2.08 & 1.22 & 1.49 & 1.83 & 1.93\\
\midrule
Running avg. 4               & 0.64 & 0.87 & 1.07 & 1.20 & 0.40 & 0.59 & 0.77 & 0.88 & 0.37 & 0.58 & 1.03 & 1.02 & 0.60 & 0.90 & 1.11 & 1.15\\
Running avg. 2               & 0.48 & 0.74 & 1.02 & 1.17 & 0.32 & 0.52 & 0.74 & 0.87 & 0.30 & 0.52 & 0.99 & 0.97 & 0.41 & 0.74 & 0.99 & 1.09\\
Zero-velocity                & 0.39 & 0.68 & 0.99 & 1.15 & 0.27 & 0.48 & 0.73 & \underline{0.86} & \bf{0.26} & \bf{0.48} & \bf{0.97} & \bf{0.95} & {\bf 0.31} & {\bf 0.67} & {\bf 0.94} & {\bf 1.04}\\
\midrule
Zero noise (SA)              & 0.44 & 0.71 & 1.16 & 1.34 & 0.39 & 0.65 & 1.13 & 1.36 & 0.51 & 0.83 & 1.48 & 1.62 & 0.57 & 1.47 & 2.08 & 2.30\\
Sampling-based loss (SA)     & 0.92 & 0.98 & 1.02 & 1.20 & 0.98 & 0.99 & 1.18 & 1.31 & 1.38 & 1.39 & 1.56 & 1.65 & 1.78 & 1.80 & 1.83 & 1.90\\
\midrule
Residual (SA)                & 0.34 & 0.60 & 0.95 & 1.09 & 0.30 & 0.53 & 0.92 & 1.13 & 0.36 & 0.66 & 1.17 & 1.27 & 0.44 & 0.93 & 1.45 & 1.60\\
Residual unsup. (MA)         & {\bf 0.27} & {\bf 0.47} & {\bf 0.70} & {\bf 0.78} & \underline{0.25} & \underline{0.43} & \underline{0.71} & 0.87 & 0.33 & 0.61 & \underline{1.04} & 1.19 & {\bf 0.31} & 0.69 & 1.03 & 1.12\\
Residual sup. (MA)           & \underline{0.28} & \underline{0.49} & \underline{0.72} & \underline{0.81} & {\bf 0.23} & {\bf 0.39} & {\bf 0.62} & {\bf 0.76} & \underline{0.33} & \underline{0.61} & 1.05 & \underline{1.15} & {\bf 0.31} & \underline{0.68} & \underline{1.01} & \underline{1.09}\\
\midrule
Untied (MA)               & 0.33 & 0.54 & 0.78 & 0.91 & 0.28 & 0.45 & 0.65 & 0.83 & 0.35 & 0.62 & 1.03 & 1.14 & 0.35 & 0.71 & 1.01 & 1.09\\
\bottomrule
\end{tabular}
\vspace{3mm}
\caption{Detailed results for motion prediction, measured in mean angle error for walking, eating, smoking and discussion activities of the Human 3.6M dataset. The top section corresponds to previous work based on deep recurrent neural networks. ``Zero noise'' is a model trained by feeding ground truth at each time step. ``Sampling-based loss'' is trained by letting the decoder feed its own output.
{\bf SA} stands for ``Single action'', and {\bf MA} stands for ``Multi-action''. Finally ``Untied'' is the same model as Residual sup (MA), but with untied weights between encoder and decoder.}
\label{tab:detailed}
\end{table*}

\begin{table*}[t]
\centering
\footnotesize
\tabcolsep=0.72mm
\begin{tabular}{@{}lrrrr|rrrr|rrrr|rrrr|rrrr|rrrr@{}}
 & \multicolumn{4}{c}{Directions} & \multicolumn{4}{c}{Greeting} & \multicolumn{4}{c}{Phoning} & \multicolumn{4}{c}{Posing} & \multicolumn{4}{c}{Purchases} & \multicolumn{4}{c}{Sitting}\\
milliseconds     & 80 & 160 & 320 & 400 & 80 & 160 & 320 & 400 & 80 & 160 & 320 & 400 & 80 & 160 & 320 & 400 & 80 & 160 & 320 & 400 & 80 & 160 & 320 & 400\\
\midrule
Zero-velocity    & {\bf 0.25} & {\bf 0.44} & {\bf 0.61} & {\bf 0.68} & 0.80 & 1.23 & 1.81 & 1.87 & 0.80 & 1.23 & 1.81 & 1.87 & {\bf0.32} & {\bf0.63} & {\bf1.16} & {\bf1.45} & 0.72 & 1.03 & 1.46 & 1.49 & \underline{0.43} & 1.12 & {\bf 1.41} & {\bf 1.58} \\
\midrule
Res. (SA)        & 0.44 & 0.95 & 1.27 & 1.55 & 0.87 & 1.40 & 2.19 & 2.26 & 0.31 & 0.57 & 0.88 & 1.04 & 0.50 & 0.96 & 1.64 & 1.96 & 0.74 & 1.60 & 1.57 & 1.72 & 0.44 & \underline{1.05} & 1.51 & 1.69\\
Res. unsup. (MA) & 0.27 & \underline{0.47} & 0.73 & 0.87 & \underline{0.77} & \underline{1.18} & {\bf 1.74} & \underline{1.84} & \underline{0.24} & {\bf 0.43} & {\bf 0.68} & \underline{0.83} & 0.40 & 0.77 & 1.32 & 1.62 & 0.62 & 1.10 & {\bf 1.07} & {\bf 1.14} & 0.68 & {\bf 1.04} & \underline{1.43} & 1.65 \\
Res. sup. (MA)   & \underline{0.26} & \underline{0.47} & \underline{0.72} & \underline{0.84} & {\bf 0.75} & {\bf 1.17} & {\bf 1.74} & {\bf 1.83} & {\bf 0.23} & {\bf 0.43} &\underline{0.69} & {\bf 0.82} & \underline{0.36} & \underline{0.71} & \underline{1.22} & \underline{1.48} & {\bf 0.51} & {\bf 0.97} & {\bf 1.07} & \underline{1.16} & {\bf 0.41} & \underline{1.05} & 1.49 & \underline{1.63}\\
\midrule
Untied (MA)   & 0.31 & 0.52 & 0.77 & 0.89 & 0.79 & 1.19 & 1.72 & {\bf 1.83} & 0.27 & 0.46 & 0.68 & 0.85 & 0.42 & 0.77 & 1.29 & 1.58 & \underline{0.52} &\underline{1.01} & {\bf 1.07} & \underline{1.16} & 0.51 & 1.13 & 1.56 & 1.74\\
\bottomrule\\
 & \multicolumn{4}{c}{Sitting down} & \multicolumn{4}{c}{Taking photo} & \multicolumn{4}{c}{Waiting} & \multicolumn{4}{c}{Walking Dog} & \multicolumn{4}{c}{Walking together} & \multicolumn{4}{c}{Average}\\
milliseconds & 80 & 160 & 320 & 400 & 80 & 160 & 320 & 400 & 80 & 160 & 320 & 400 & 80 & 160 & 320 & 400 & 80 & 160 & 320 & 400 & 80 & 160 & 320 & 400 \\
\midrule
Zero-velocity    & {\bf 0.27} & {\bf 0.54} & {\bf 0.93} & {\bf 1.05} & {\bf 0.22} & {\bf 0.47} & {\bf 0.78} & {\bf 0.89} & {\bf 0.27} & {\bf 0.49} & {\bf 0.96} & {\bf 1.12} & 0.60 & 0.96 & 1.27 & \underline{1.33} & \underline{0.33} & \underline{0.60} & 0.96 & 1.03 & 0.42 & 0.74 & 1.12 & \underline{1.20} \\
\midrule
Res. (SA)        & \underline{0.38} & \underline{0.77} & \underline{1.36} & \underline{1.59} & 0.37 & 0.66 & 1.30 & 1.70 & 0.36 & 0.73 & 1.31 & 1.51 & 0.62 & 1.02 & 1.55 & 1.65 & 0.44 & 0.81 & 1.25 & 1.36 & 0.46 & 0.88 & 1.35 & 1.54 \\
Res. unsup. (MA) & 0.41 & 0.80 & 1.43 & 1.63 & 0.27 & 0.56 & 0.98 & 1.16 & 0.32 & 0.62 & 1.13 & 1.30 & \underline{0.58} & \underline{0.95} & \underline{1.37} & 1.45 & 0.35 & 0.62 & \underline{0.87} & {\bf 0.87} & \underline{0.39} & \underline{0.72} & \underline{1.08} & 1.22 \\
Res. sup. (MA)   & 0.39 & 0.81 & 1.40 & 1.62 & \underline{0.24} & \underline{0.51} & \underline{0.90} & \underline{1.05} & \underline{0.28} & \underline{0.53} & \underline{1.02} & \underline{1.14} & {\bf 0.56} & {\bf 0.91} & {\bf 1.26} & {\bf 1.40} & {\bf 0.31} & {\bf 0.58} & {\bf 0.87} & \underline{0.91} & {\bf 0.36} & {\bf 0.67} & {\bf 1.02} & {\bf 1.15} \\
\midrule
Untied (MA)   & 0.47 & 0.89 & 1.57 & 1.72 & 0.30 & 0.56 & 0.95 & 1.12 & 0.38 & 0.64 & 1.18 & 1.41 & 0.61 & 0.98 & 1.42 & 1.54 & 0.40 & 0.69 & 0.98 & 1.03 & 0.42 & 0.74 & 1.11 & 1.26  \\
\bottomrule
\end{tabular}
\vspace{3mm}
\caption{Prediction results for our zero-velocity baseline and our main prediction methods on the remainder 11 actions of the H3.6m dataset.}
\label{tab:detailed2}
\vspace{-3mm}
\end{table*}

\vspace{-3mm}
\paragraph{Dataset and data pre-processing.} Following previous work, we use the Human 3.6M (H3.6M) dataset by Ionescu~\etal~\cite{h36m}, which is currently the largest publicly available dataset of motion capture data. H3.6M includes seven actors performing 15 varied activities such as walking, smoking, engaging in a discussion, taking pictures, and talking on the phone, each in two different trials.
For a fair comparison, we adopt the pose representation and evaluation loss from~\cite{fragkiadaki,srnn}.
Pose is represented as an exponential map representation of each joint, with a special pre-processing of global translation and rotation (see ~\cite{crbm} for more details).
For evaluation, similar to~\cite{fragkiadaki,srnn}, we measure the Euclidean distance between our prediction and the ground truth in angle-space for increasing time horizons. We report the average error on eight randomly sampled test sequences, and use the sequences of subject five for testing, while the rest of the sequences are used for training.

\vspace{-3mm}
\paragraph{A scalable seq2seq architecture.} In all our experiments, we use a single gated recurrent unit~\cite{gru} (GRU) with 1024 units, as a computationally less-expensive alternative to LSTMs, and we do not use any time-independent layers for representation learning. Experimentally, we found that stacking recurrent layers makes the architecture harder to train, while it also makes it slower; we also found that the best performance is obtained without a spatial encoder. We do, however, use a spatial decoder to back-project the 1024-dimensional output of the GRU to 54 dimensions, the number of independent joint angles provided in H3.6M.

We use a learning rate of 0.005 in our multi-action experiments, and a rate of 0.05 in our action-specific experiments; in both cases, the batch size is 16, and we clip the gradients to a maximum L2-norm of 5. During training as well as testing, we feed 2 seconds of motion to the encoder, and predict either 1 second (for long-term experiments) or 400 milliseconds (for short-term prediction) of motion from the decoder.
We implemented our architecture using TensorFlow~\cite{tensorflow}, which takes 75ms for forward processing and back-propagation per iteration on an NVIDIA Titan GPU.

\vspace{-3mm}
\paragraph{Baselines.}
We compare against two recent approaches to human motion prediction based on deep RNNs: LSTM-3LR and ERD by Fragkiadaki~\etal~\cite{fragkiadaki}, and SRNN by Jain~\etal~\cite{srnn}. To reproduce previous work, we rely on the pre-trained models and implementations of ERD, LSTM-3LR and SRNN publicly available \footnote{\url{https://github.com/asheshjain399/RNNexp}}.
These implementations represent the best efforts of the SRNN authors to reproduce the results of the ERD and LSTM-3LR models reported by Fragkiadaki~\etal~\cite{fragkiadaki}, as there is no official public implementation for that work.
We found that, out of the box, these baselines produce results slightly different (most often better) from those reported by Jain~\etal~\cite{srnn}.

We also consider an agnostic zero-velocity baseline which constantly predicts the last observed frame. For completeness, we also consider running averages of the last two and four observed frames. While these baselines are very simple to implement, they have not been considered in recent work that uses RNNs to model human motion.

\begin{figure*}[t!]
  \includegraphics[width=\linewidth,trim=15mm 60mm 15mm 40mm,clip=true]{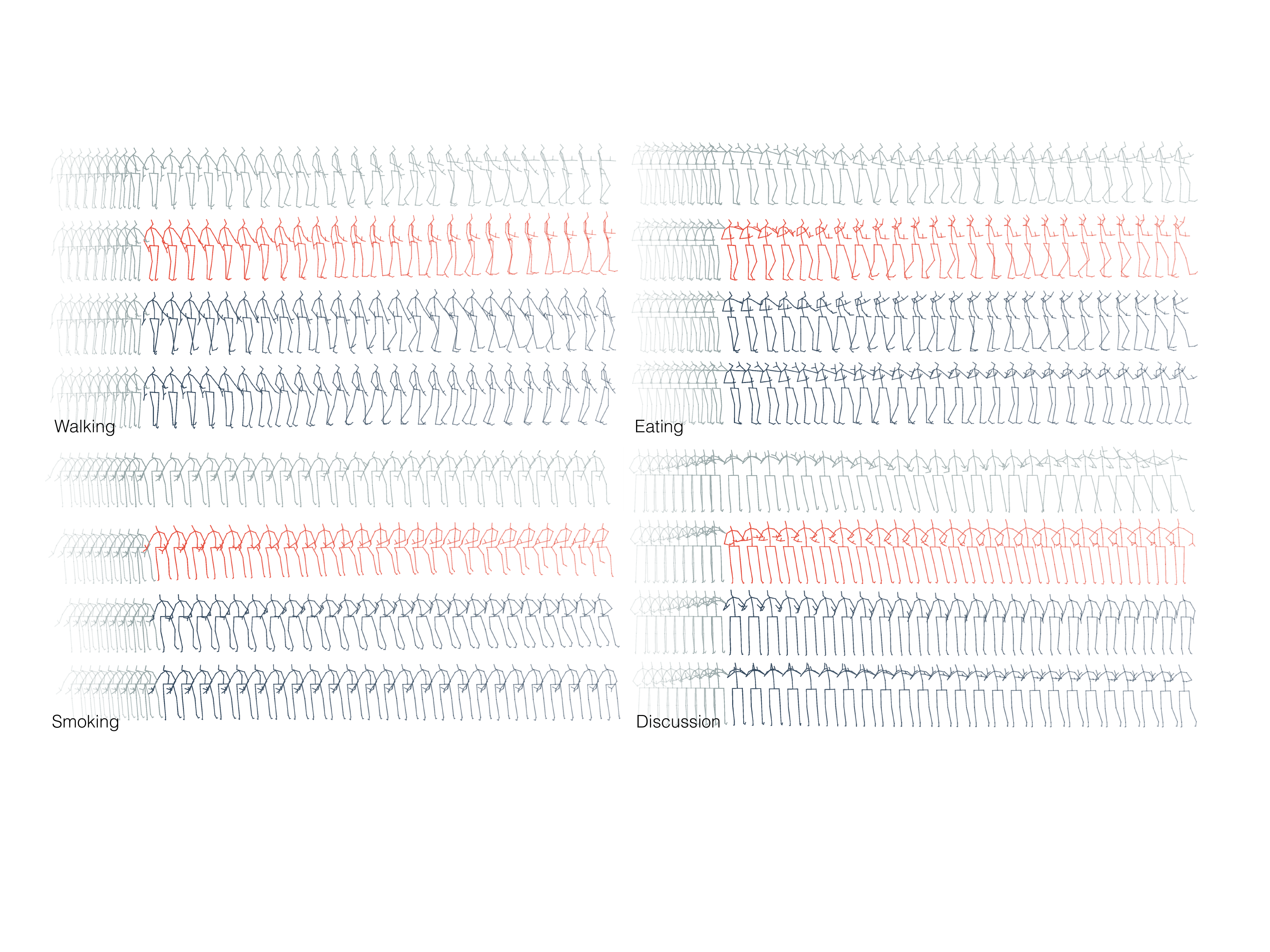}
  \caption{
  Qualitative long-term motion generation, showing two seconds of motion prediction on different activities. The gray top sequence corresponds to ground truth and the red one to SRNN. The first dark blue sequence corresponds to our method, trained on specific actions, and without residual connection, but using sampling-based loss ({\bf Sampling-based loss (SA)} on Table~\ref{tab:detailed}). This model produces plausible motion in the long term, but does suffer from discontinuities in short-term predictions. The last blue sequence corresponds to our full model, including residual connections, and trained on multiple actions ({\bf Residual sup. (MA)} on Table~\ref{tab:detailed}); this model produces smooth, continuous predictions in the short term, but converges to a mean pose.}
  \label{fig:qualitative}
  \vspace{-3mm}
\end{figure*}

\section{Results}

Figure~\ref{fig:losses} shows a summary of the results obtained by ERD, LSTM-3LR and SRNN, as well as a zero-velocity baseline and our method,
on four actions of the Human 3.6 dataset.
Tables~\ref{tab:detailed} and~\ref{tab:detailed2} describe these results in more detail, and include results on the rest of the actions.
In the remainder of the section we analyze these results.

\vspace{-3mm}
\paragraph{Zero-velocity baseline.} The first striking result is the comparatively good performance of the baselines, specially the zero-velocity one. They clearly outperform state-of-the-art results, highlighting the severity of the discontinuities between conditioning and prediction in previous work. The good performance of the baseline also means that deterministic losses are not suitable to evaluate motion forecasting with a long time horizon.

\vspace{-3mm}
\paragraph{Sampling-based loss.}
In Table~\ref{tab:detailed}, 
using our sampling-based loss consistently achieves motion prediction error competitive with or better than the state of the art.
Moreover, since we have trained our model to minimize the error over a 1-second time horizon, the network retains the ability to generate plausible motion in the long run. Figure~\ref{fig:qualitative} shows a few qualitative examples of long-term motion using this approach.
Given that our proposed sampling-based loss does not require any hyper-parameter tuning, we would argue that this is a fast-to-train, interesting alternative to previous work for long-term motion generation using RNNs.

\vspace{-3mm}
\paragraph{Residual architecture and multi-action models.}

Finally, we report the performance obtained by our architecture with sampling-based loss, residual connections and trained on single (SA) or multiple actions (MA) in the bottom subgroup of Table~\ref{tab:detailed}.
We can see that using a residual connection greatly improves performance and pushes our method beyond the state of the art, which highlights the fact that velocity representations are easier to model by our network.
Importantly, our method obtains its best performance 
when trained on multiple actions;
this result, together with the simplicity of our approach, uncovers the importance of large amounts of training data when learning short-term motion dynamics.
We also note that highly aperiodic classes such as discussion, directions and sitting down remain very hard to model.

Moreover, we observe that adding semantic information to the network in the form of action labels helps in most cases, albeit by a small margin. Likely, this is due to the fact that, for short-term motion prediction, modelling physical constraints (\eg momentum preservation) is more important than modelling high-level semantic intentions.

When analysing Fig.~\ref{fig:qualitative}, it becomes obvious that the best numerical results do not correspond to the best qualitative long-term motion -- a result that persists even when trained to minimize loss over long horizons (\eg 1 second). One can hardly blame the method though, since our network is achieving the lowest loss in an independent validation set. In other words, the network is excelling in the task that has been assigned to it. 
In order to produce better qualitative results, we argue that a different loss that encourages other similarity measures (\eg adversarial, entropy-based \etc) should be used instead. Our results suggest that it is inherently hard to produce both accurate short-term predictions -- which are relatively deterministic and seem to be properly optimized with the current loss -- and long-term forecasting using RNNs.

\section{Conclusions and future work}

We have demonstrated that previous work on human motion modelling using deep RNNs has harshly neglected the important task of short-term motion prediction, as we have shown that a zero-velocity prediction is a simple but hard-to-beat baseline that largely outperforms the state of the art. Based on this observation, we have developed a sequence-to-sequence architecture with residual connections which, when trained on a sample-based loss, outperforms previous work. Our proposed architecture, being simple and scalable, can be trained on large-scale datasets of human motion,
which we have found to be crucial to learn the short-term dynamics of human motion. Finally, we have shown that providing high-level supervision to the network in the form of action labels improves performance, but an unsupervised baseline is very competitive nonetheless.
We find this last result particularly encouraging, as it departs from previous work in human motion modelling which has typically worked on small, action-specific datasets. Future work may focus on exploring ways to use even larger datasets of motion capture in an unsupervised manner.

\vspace{-5mm}
\paragraph{Acknowledgements.}
The authors thank Laura Sevilla-Lara for proofreading our work.
We also thank Nvidia for the donation of some of the GPUs used in this research.
This research was supported in part by NSERC.

{\small
\bibliographystyle{ieee}
\bibliography{egbib}
}

\end{document}